\def\year{2022}\relax
\documentclass[letterpaper]{article} % DO NOT CHANGE THIS
\usepackage{aaai22}  % DO NOT CHANGE THIS
\usepackage{times}  % DO NOT CHANGE THIS
\usepackage{helvet}  % DO NOT CHANGE THIS
\usepackage{courier}  % DO NOT CHANGE THIS
\usepackage[hyphens]{url}  % DO NOT CHANGE THIS
\usepackage{graphicx} % DO NOT CHANGE THIS
\usepackage{natbib}  % DO NOT CHANGE THIS AND DO NOT ADD ANY OPTIONS TO IT
\usepackage{caption} % DO NOT CHANGE THIS AND DO NOT ADD ANY OPTIONS TO IT
\DeclareCaptionStyle{ruled}{labelfont=normalfont,labelsep=colon,strut=off} % DO NOT CHANGE THIS
\usepackage{algorithm}
\usepackage{algorithmic}

\usepackage{microtype}
\usepackage{graphicx}
\usepackage{caption}
\usepackage{subcaption}
\usepackage{booktabs} % for professional tables
\usepackage{tikz-cd}
\usetikzlibrary{arrows}
\usepackage{amsmath}
\usepackage{comment}
\usepackage{amsthm}
\usepackage{amssymb}
\usepackage{url}
\usepackage{tikz}
\usepackage{tikz-qtree,tikz-qtree-compat}
\usepackage{soul}

\makeatletter
\newtheorem*{rep@theorem}{\rep@title}
\newcommand{\newreptheorem}[2]{%
\newenvironment{rep#1}[1]{%
 \def\rep@title{#2 \ref{##1}}%
 \begin{rep@theorem}}%
 {\end{rep@theorem}}}
\makeatother

\newtheorem{theorem}{Theorem}
\newreptheorem{theorem}{Theorem}

\usetikzlibrary{positioning}

\newtheorem{lemma}[theorem]{Lemma}

\newcommand\independent{\protect\mathpalette{\protect\independenT}{\perp}}
\def\independenT#1#2{\mathrel{\rlap{$#1#2$}\mkern2mu{#1#2}}}

\usepackage{newfloat}
\usepackage{listings}
\floatstyle{ruled}
\newfloat{listing}{tb}{lst}{}
\floatname{listing}{Listing}
\title{Unit Selection with Causal Diagram}
\author {
    Ang Li,\textsuperscript{\rm 1}
    Judea Pearl, \textsuperscript{\rm 1}
}
\affiliations {
    \textsuperscript{\rm 1} Department of Computer Science, University of California, Los Angeles, California, USA\\
    angli@cs.ucla.edu, judea@cs.ucla.edu
}

\begin{document}

\maketitle

\begin{abstract}
The unit selection problem aims to identify a set of individuals who are most likely to exhibit a desired mode of behavior, for example, selecting individuals who would respond one way if encouraged and a different way if not encouraged. Using a combination of experimental and observational data, Li and Pearl derived tight bounds on the ``benefit function" - the payoff/cost associated with selecting an individual with given characteristics. This paper shows that these bounds can be narrowed significantly (enough to change decisions) when structural information is available in the form of a causal model. We address the problem of estimating the benefit function using observational and experimental data when specific graphical criteria are assumed to hold.
\end{abstract}
\section{Introduction}\label{sec:intro}
In many areas of industry, marketing, and health science, the unit selection dilemma arises. For example, in customer relationship management \cite{berson1999building, lejeune2001measuring, hung2006applying, tsai2009customer}, it is useful to know which customers are going to churn but might reconsider if encouraged to stay. Due to the high expense of such initiatives, management is forced to limit inducement to customers who are most likely to exhibit the behavior of interest. As another example, companies are interested in identifying users who would click on an advertisement if and only if it is highlighted in online advertising \cite{yan2009much, bottou2013counterfactual, li2014counterfactual, sun2015causal}. The challenge in identifying these users stems from the fact that the desired response pattern is not observed directly but rather is defined counterfactually in terms of what the individual would do under hypothetical unrealized conditions. For example, when we observe that a user has clicked on a highlighted advertisement, we do not know whether they would click on that same advertisement if it were not highlighted. 

The benefit function for the unit selection problem was defined by Li and Pearl \cite{li2019unit}, and it properly captures the nature of the desired behavior. Using a combination of experimental and observational data, Li and Pearl derived tight bounds of the benefit function. The only assumption is that the treatment has no effect on the population specific characteristics. However, Li-Pearl's derivation does not leverage
information from auxiliary covariates, if such is available. Mueller, Li, and Pearl \cite{mueller2021causes} recently proposed using covariate information and the causal structure to narrow the bounds of probability of necessity and sufficiency. Dawid et al. \cite{dawid2017} also proposed using covariates information to narrow the bounds of probability of necessity. A similar approach might be used for the benefit function. Most crucially, the information provided by covariates and their causal structure may result in a reversal of decision (relative to not considering such covariates).

Consider the following motivating scenario: a carwash company wants to offer a discount to employees of company A. The offer can only be presented to the entire company A; the carwash company will not be able to provide a discount to a specific group inside the company A. The carwash company's manager seeks to maximize total profit, including nonimmediate profit. The management estimates that the benefit of selecting a complier (i.e., offer the discount to a customer who would use the carwash service if they received the discount, but would not otherwise) is $\$100$ as the profit is $\$140$ but the discount is $\$40$, that of selecting an always-taker (i.e., offer the discount to a customer who would use the carwash service regardless of whether they received the discount) is $-\$60$ as the customer would use the service anyway (so the company loses the value of the discount and an extra cost of $\$20$ because the always-taker may require additional discounts in the future), that of selecting a never-taker (i.e., offer the discount to a customer who would never use the carwash service regardless of whether they received the discount) is $\$0$ as the cost of issuing the discount is negligible, and that of selecting a defier (i.e., offer the discount to a customer who would not use the carwash service if they received the discount, but would use the carwash service otherwise) is $-\$140$ as the customer is lost due to the discount. The manager of carwash company has both experimental and observational data related to customer age collected from the company A. If the entire company A's employees are given the discount, the manager of carwash wants to know what the average profit will be.

Based on Li-Pearl's model, it is easy to see that the benefit vector for the aforementioned example is $(100, -60, 0, -140)$, and a corresponding benefit function can be defined as the objective function. Li-Pearl's model can then obtain the bounds of the benefit function using experimental and observational data. The model, however, does not take into account the covariate information (customer age) and the causal structure.

In this paper, we show how the information included in such covariates and their causal structure, can be used to narrow the bounds of the benefit function in Li-Pearl's model. Most importantly, the narrower bounds can, sometimes, flip the decision.

\section{Preliminaries}
\label{related work}
In this section, we review Li and Pearl's benefit function of the unit selection problem \cite{li2019unit}. Individual behavior was classified into four response types: labeled complier, always-taker, never-taker, and defier. Suppose the benefit of selecting one individual in each category are $\beta, \gamma, \theta, \delta$ respectively (i.e., the benefit vector is $(\beta, \gamma, \theta, \delta)$). They defined the objective function of the unit selection problem as the average benefit gained per individual. Suppose $a$ and $a'$ are binary treatments, $r$ and $r'$ are binary outcomes, and $c$ are population-specific characteristics, the objective function (i.e., benefit function) is following (If the goal is to evaluate the average benefit gained per individual for a specific population $c$, $argmax_c$ can be dropped.):
\begin{eqnarray*}
\label{liobj}
argmax_c \text{ }\beta P(r_{a},r'_{a'}|c)+\gamma P(r_{a},r_{a'}|c) + \nonumber \\+\theta P(r'_{a},r'_{a'}|c)+\delta P(r'_{a},r_{a'}|c).
\end{eqnarray*}
Using a combination of experimental and observational data, Li and Pearl established the most general tight bounds on this benefit function (which we refer to as Li-Pearl's Theorem in the rest of the paper). The only constraint is that the population-specific characteristics are not a descendant of the treatment.

However, the information of covariates (if available, such as the age in the motivating example in the previous section) are not considered. In this paper, we present three common cases of covariates and their causal structures and theorems that show how the information about the covariates along with their causal structures could narrow the bounds of the benefit function. The improvement of the bounds is sometimes significant and can change the decisions compared to Li-Pearl's Theorem.

\section{Selection Criteria with Causal Diagrams}
We present three common cases of the covariates and their causal structures in this section. For each case, we provide a theorem for estimating the benefit function in such a case. The proof of all theorems is in the appendix. In any causal diagram of this paper, the dot line between $A$ and $B$ represents either $A$ affects $B$, $B$ affects $A$, or $A$ and $B$ are independent; the dot line with arrow from $A$ to $B$ represents either $A$ affects $B$ or $A$ and $B$ are independent.
\subsection{Causal Diagram with Non-descendant Covariates}
%\subsubsection{Non-descendant $Z\cup C$}
Theorem \ref{thm1} provides bounds for the benefit function when a set $Z$ of variables can be measured, which satisfies only one condition: both population-specific variables $C$ and covariates $Z$ contain no descendant of $X$. This condition is important because if $X$ is set to $x$ and $C\cup Z$ contains a descendant of $X$, then $C\cup Z$ could be altered and $P(y_x|z,c)$ would be another unmeasurable counterfactual term. If the descendant is independent of $Y_x$, then $P(y_x|z,c)$ would be measurable, but the descendant would not contribute to any narrowing of the bounds. These bounds are always contained within the bounds of the benefit function in Li-Pearl's Theorem.

\begin{theorem}
\label{thm1}
Given a causal diagram $G$ and distribution compatible with $G$, let $Z\cup C$ be a set of variables that does not contain any descendant of $X$ in $G$, then the benefit function $f(c)=\beta P(y_x,y'_{x'}|c)+\gamma P(y_x,y_{x'}|c)+ \theta P(y'_x,y'_{x'}|c) + \delta P(y_{x'},y'_{x}|c)$ is bounded as follows:
\begin{eqnarray*}
W+\sigma U\le f(c) \le W+\sigma L\text{~~~~~~~~if }\sigma < 0,\\
W+\sigma L\le f(c) \le W+\sigma U\text{~~~~~~~~if }\sigma > 0,
\end{eqnarray*}
where $\sigma, W,L,U$ are given by,
\begin{eqnarray*}
&&\sigma = \beta - \gamma - \theta + \delta,\\
&&W=(\gamma -\delta)P(y_x|c)+\delta P(y_{x'}|c)+\theta P(y'_{x'}|c),\\
&&L=\sum_z\max\left\{
\begin{array}{c}
0,\\
P(y_x|z,c)-P(y_{x'}|z,c),\\
P(y|z,c)-P(y_{x'}|z,c),\\
P(y_x|z,c)-P(y|z,c)
\end{array}
\right\}\\ &&\times P(z|c),\\
&&U=\sum_z\min\left\{
\begin{array}{c}
P(y_x|z,c),\\
P(y'_{x'}|z,c),\\
P(y,x|z,c)+P(y',x'|z,c),\\
P(y_x|z,c)-P(y_{x'}|z,c)+\\
+P(y,x'|z,c)+P(y',x|z,c)
\end{array}
\right\}\\ &&\times P(z|c).
\end{eqnarray*}
\end{theorem}

Notably, $C$ can be interpreted as the population-specific variables, and $Z$ are the attributes in each population. Moreover, the bounds provided above are always no worse than Li-Pearl's bound (see proof in the appendix). Besides, if $\sigma=0$, the Gain Equality is satisfied in Li-Pearl's model, and the result of the benefit function is no longer bounds, but a point estimate. 

\subsection{Causal Diagram with Mediators}
\subsubsection{Partial Mediators}
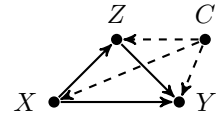
\begin{figure}[h]
\centering
\begin{tikzpicture}[->,>=stealth',node distance=2cm,
  thick,main node/.style={circle,fill,inner sep=1.5pt}]
%  \node[main node] (0) [label=above:{$U$}]{};
  \node[main node] (1) [label=above:{$Z$}]{};
  \node[main node] (2) [below left =1cm of 1,label=left:$X$]{};
  \node[main node] (3) [below right =1cm of 1,label=right:$Y$] {};
  \node[main node] (4) [right =1cm of 1,label=above:{$C$}]{};
  \path[every node/.style={font=\sffamily\small}]
    (2) edge node {} (1)
    (1) edge node {} (3)
    (2) edge node {} (3);
%  \draw [dashed] (0) -- (1);
%  \draw [dashed] (0) -- (2);
  \draw [dashed, ->] (4) -- (2);
  \draw [dashed, ->] (4) -- (3);
  \draw [dashed, ->] (4) -- (1);
\end{tikzpicture}
\caption{Mediator $Z$ with direct effects of $X$ on $Y$.}
% \caption{Mediator $Z$ with direct effect and confounder}
\label{causalg2}
\end{figure}
In Figure \ref{causalg2}, partial mediator $Z$ is a descendant of $X$; thus, we cannot use Theorem \ref{thm1}. However, the absence of confounders (other than population specific variables $C$) between $Z$ and $Y$ and between $X$ and $Y$ permits us to bound the benefit function as follows:
\vspace{10pt}
\begin{theorem}
\label{thm2}
Given a causal diagram $G$ and distribution compatible with $G$, let $Z$ be a set of variables such that $\forall x,x' \in X : x \ne x', (Y_x \independent X\cup Z_{x'}\ |\ Z_x, C)$ in $G$, and $C$ does not contain any descendant of $X$ in $G$, then the benefit function $f(c)=\beta P(y_x,y'_{x'}|c)+\gamma P(y_x,y_{x'}|c)+ \theta P(y'_x,y'_{x'}|c) + \delta P(y_{x'},y'_{x}|c)$ is bounded as follows:
\begin{eqnarray*}
W+\sigma U\le f(c) \le W+\sigma L\text{~~~~~~~~if }\sigma < 0,\\
W+\sigma L\le f(c) \le W+\sigma U\text{~~~~~~~~if }\sigma > 0,
\end{eqnarray*}
where $\sigma, W,L,U$ are given by,
\begin{eqnarray*}
&&\sigma = \beta - \gamma - \theta + \delta,\\
&&W=(\gamma -\delta)P(y_x|c)+\delta P(y_{x'}|c)+\theta P(y'_{x'}|c),\\
&&L=\max\left\{
\begin{array}{c}
0,\\
P(y_x|c)-P(y_{x'}|c),\\
P(y|c)-P(y_{x'}|c),\\
P(y_x|c)-P(y|c)\\
\end{array}
\right\},\\
&&U=\min\left\{
\begin{array}{c}
P(y_x|c),\\
P(y'_{x'}|c),\\
P(y,x|c)+P(y',x'|c),\\
P(y_x|c)-P(y_{x'}|c)+\\+P(y,x'|c)+P(y',x|c),\\
\sum_z \sum_{z'} \min\{P(y|z,x,c),\\P(y'|z',x',c)\}\times\\ \min\{P(z_x|c),P(z'_{x'}|c)\}
\end{array}
\right\}.
\end{eqnarray*}
\end{theorem}

Although this lower bound is unchanged from that in Li-Pearl's Theorem, the upper bound contains a vital additional argument (i.e., the last term in the min function of $U$) to the min function. This new term can significantly reduce the upper bound. The rest of the terms are included because sometimes the bounds of Li-Pearl's Theorem are superior. The following theorem has the same quality.

\subsubsection{Pure Mediators}
Figure \ref{causalg3} is a special case of Figure \ref{causalg2}, in which $X$ has no direct effects on $Y$. The resulting bounds for the benefit function are as follows:
\vspace{10pt}
\begin{theorem}
\label{thm3}
Given a causal diagram $G$ in Figure \ref{causalg3} and distribution compatible with $G$, and $C$ does not contain any descendant of $X$, then the benefit function $f(c)=\beta P(y_x,y'_{x'}|c)+\gamma P(y_x,y_{x'}|c)+ \theta P(y'_x,y'_{x'}|c) + \delta P(y_{x'},y'_{x}|c)$ is bounded as follows:
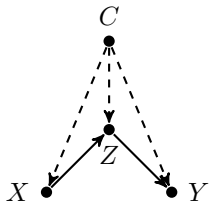
\begin{figure}[h]
\centering
\begin{tikzpicture}[->,>=stealth',node distance=2cm,
  thick,main node/.style={circle,fill,inner sep=1.5pt}]
  \node[main node] (0) [label=above:{$C$}]{};
  \node[main node] (1) [below =1cm of 0,label=below:{$Z$}]{};
  \node[main node] (2) [below left =1cm of 1,label=left:$X$]{};
  \node[main node] (3) [below right =1cm of 1,label=right:$Y$] {};
  \path[every node/.style={font=\sffamily\small}]
    (2) edge node {} (1)
    (1) edge node {} (3);
  \draw [dashed, ->] (0) -- (2);
  \draw [dashed, ->] (0) -- (3);
  \draw [dashed, ->] (0) -- (1);
\end{tikzpicture}
\caption{Mediator $Z$ with no direct effects of $X$ on $Y$.}
\label{causalg3}
\end{figure}
\begin{eqnarray*}
W+\sigma U\le f(c) \le W+\sigma L\text{~~~~~~~~if }\sigma < 0,\\
W+\sigma L\le f(c) \le W+\sigma U\text{~~~~~~~~if }\sigma > 0,
\end{eqnarray*}
where $\sigma, W,L,U$ are given by,
\begin{eqnarray*}
&&\sigma = \beta - \gamma - \theta + \delta,\\
&&W=(\gamma -\delta)P(y_x|c)+\delta P(y_{x'}|c)+\theta P(y'_{x'}|c),\\
&&L=\max\left\{
\begin{array}{c}
0,\\
P(y_x|c)-P(y_{x'}|c),\\
P(y|c)-P(y_{x'}|c),\\
P(y_x|c)-P(y|c)\\
\end{array}
\right\},\\
&&U=\min\left\{
\begin{array}{c}
P(y_x|c),\\
P(y'_{x'}|c),\\
P(y,x|c)+P(y',x'|c),\\
P(y_x|c)-P(y_{x'}|c)+\\+P(y,x'|c)+P(y',x|c),\\
\Sigma_z \Sigma_{z'\ne z}\min\{P(y|z,c),\\P(y'|z',c)\}\times\\
\min\{P(z|x,c),P(z'|x',c)\}
\end{array}
\right\}.
\end{eqnarray*}
\end{theorem}

The core term (i.e., the last term in the min function of $U$) for Theorem \ref{thm3} added to the upper bound notably only requires observational data.

\section{Examples}\label{example}
In this section, we will show how the presented theorems can be applied to applications and how the theorems affect judgments using two cases.

\subsection{Company Selection}
Consider the motivating example in the introduction section.

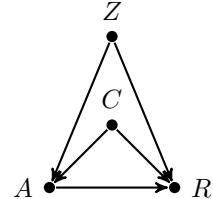
\begin{figure}[h]
\centering
\begin{tikzpicture}[->,>=stealth',node distance=2cm,
  thick,main node/.style={circle,fill,inner sep=1.5pt}]
%  \node[main node] (4) [label=above:{$U$}]{};
  \node[main node] (0) [label=above:{$Z$}]{};
  \node[main node] (1) [below =1cm of 0,label=above:{$C$}]{};
  \node[main node] (2) [below left =1cm of 1,label=left:$A$]{};
  \node[main node] (3) [below right =1cm of 1,label=right:$R$] {};
  \path[every node/.style={font=\sffamily\small}]
    (0) edge node {} (2)
    (0) edge node {} (3)
    (2) edge node {} (3)
    (1) edge node {} (2)
    (1) edge node {} (3);
%   \draw [dashed, -] (4) -- (2);
%   \draw [dashed, -] (4) -- (3);
%   \draw [dashed, -] (1) -- (3);
\end{tikzpicture}
\caption{Company selection model.}
\label{causalg1}
\end{figure}

Let $A=a$ denote the event that a customer receives the discount, $A=a'$ denote the event that a customer does not receive the discount, $R=r$ denote the event that a customer uses the services, $R=r'$ denote the event that a customer does not use the services, $C=c$ denote a company A's customer, $Z=z$ denote a younger customer (age below or equal to 50), and $Z=z'$ denote an older customer (age above 50). The model is as shown in Figure \ref{causalg1}.

Based on Li-Pearl's model, it is easy to see that the benefit vector is $(100, -60, 0, -140)$ (see the introduction section). Therefore, the benefit function is:
\begin{eqnarray}
argmax_c \text{ }100 P(r_{a},r'_{a'}|c)-60 P(r_{a},r_{a'}|c) + \nonumber\\+0 P(r'_{a},r'_{a'}|c)-140 P(r'_{a},r_{a'}|c).
\label{obj1}
\end{eqnarray}

The manager of the carwash company collected the data listed in Tables \ref{tb1} and \ref{tb2} from company A. By Li-Pearl's Theorem, the bounds of the benefit function are $[-0.423, 2.832]$ (see the appendix for details), and the midpoint is $1.205$. It suggests that the carwash company would gain $\$1.205$ profit from each individual from company A if they offer company A's employees the discount. Besides, most of the bounded area is positive, which provided more confidence that the conclusion is correct. However, Li-Pearl's theorem only uses the overall data in Tables \ref{tb1} and \ref{tb2} (i.e., customer age is not considered).

\begin{table}
\centering
\caption{Experimental data collected by the carwash company. 350 customers were forced to receive the discount and 350 customers were forced not to receive the discount.}
\begin{tabular}{|c|c|c|}
\hline 
&Discount&No Discount\\
\hline
Young&\begin{tabular}{c}$45$ out of $101$ \\used the service \\($44.6\%$)\end{tabular}&\begin{tabular}{c}$5$ out of $101$ \\used the service \\($5.0\%$)\end{tabular}\\
\hline
Elder&\begin{tabular}{c}$248$ out of $249$ \\used the service \\($99.6\%$)\end{tabular}&\begin{tabular}{c}$179$ out of $249$ \\used the service \\($71.9\%$)\end{tabular}\\
\hline
Overall&\begin{tabular}{c}$293$ out of $350$ \\used the service \\($83.7\%$)\end{tabular}&\begin{tabular}{c}$184$ out of $350$ \\used the service \\($52.6\%$)\end{tabular}\\
\hline
\end{tabular}
\label{tb1}
\end{table}

\begin{table}
\centering
\caption{Observational data collected by the carwash company. 700 customers were given access to the discount, they can choose whether to obtain the discount by themselves (note that a customer may still not use the service even they obtained the discount by themselves).}
\begin{tabular}{|c|c|c|}
\hline 
&Discount&No Discount\\
\hline
Young&\begin{tabular}{c}$90$ out of $152$ \\used the service \\($59.2\%$)\end{tabular}&\begin{tabular}{c}$9$ out of $50$ \\used the service \\($18.0\%$)\end{tabular}\\
\hline
Elder&\begin{tabular}{c}$157$ out of $159$ \\used the service \\($98.7\%$)\end{tabular}&\begin{tabular}{c}$239$ out of $339$ \\used the service \\($70.5\%$)\end{tabular}\\
\hline
Overall&\begin{tabular}{c}$247$ out of $311$ \\used the service \\($79.4\%$)\end{tabular}&\begin{tabular}{c}$248$ out of $389$ \\used the service \\($63.8\%$)\end{tabular}\\
\hline
\end{tabular}
\label{tb2}
\end{table}

Now, if we apply Theorem \ref{thm1} to the data in Tables \ref{tb1} and \ref{tb2}, the bounds of the benefit function is $[-0.168,-0.077]$ (see the appendix for details), with the midpoint at $-0.123$. This suggests that if the carwash company offers the discount to company A's employees, the carwash company will lose $\$0.123$ profit per individual. Notably, the upper bound ($-0.077$) is negative, implying that the carwash company must lose profit if they offers the discount to company A's employees regardless of how the bounds are used.

\subsection{Effective Patients of a Drug}
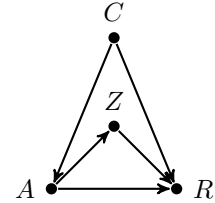
\begin{figure}
            \centering
            \begin{tikzpicture}[->,>=stealth',node distance=2cm,
              thick,main node/.style={circle,fill,inner sep=1.5pt}]
              \node[main node] (0) [label=above:{$C$}]{};
              \node[main node] (1) [below =1cm of 0,label=above:{$Z$}]{};
              \node[main node] (3) [below left =1cm of 1,label=left:$A$]{};
              \node[main node] (4) [below right =1cm of 1,label=right:$R$] {};
              \path[every node/.style={font=\sffamily\small}]
                (0) edge node {} (3)
                (0) edge node {} (4)
                (3) edge node {} (1)
                (1) edge node {} (4)
                (3) edge node {} (4);
            \end{tikzpicture}
            \caption{A graphical model representing the effects of a new drug, with $A$ representing drug usage, $R$ representing recovery, $Z$ representing blood pressure (measured at the end of the study), and $C$ representing the population specific variables (gender and age).}
            \label{unit2}
        \end{figure}

When a pharmaceutical company develops a new drug, it seeks to identify patients so as to maximize the difference between the number of effective patients and the number of ineffective patients. The causal diagram is shown in Figure \ref{unit2}. 

For the benefit vector, the pharmaceutical company assigned $1$ to a complier because the complier is the patient cured by the drug, assign $-1$ to an always-taker, a never-taker, and a defier because they are all ineffective patients. The benefit vector is then $(1,-1,-1,-1)$.

Let $A=a$ denote the event that a patient takes the drug, $A=a'$ denote the event that a patient does not take the drug, $R=r$ denote the event that a patient is recovered, $R=r'$ denote the event that a patient is not recovered, $Z=z$ denote low blood pressure (measured at the end of the study), $Z=z'$ denote high blood pressure, and $C$ (a set of variables) denote the population-specific characteristics (gender and age) of a patient. The benefit function is then
\begin{eqnarray}
argmax_c \text{ } P(r_{a},r'_{a'}|c) - P(r_{a},r_{a'}|c) -\nonumber\\- P(r'_{a},r'_{a'}|c) - P(r'_{a},r_{a'}|c).
\label{obj2}
\end{eqnarray}

The pharmaceutical company records the recovery rates of 70000 patients who were given access to the drug (i.e., observational study). For each group of patients who have the same gender and age, they record the number of patients who chose to take the drug and their recovery rates, the number of patients who did not choose to take the drug, and their recovery rates. For example, the results of the 30 years old male patients ($1075$ patients) are shown in Table \ref{tb3}.

\begin{table}[]
    \centering
\caption{Results of an observational study (30 years old male) into a new drug, with post-treatment blood pressure taken into account.}
            \begin{tabular}{|c|c|c|}
            \hline 
            &Drug&No Drug\\
            \hline
            \begin{tabular}{c}Low\\BP\end{tabular}&\begin{tabular}{c}$375$ out of $405$ \\recovered\\ ($92.6\%$)\end{tabular}&\begin{tabular}{c}$159$ out of $481$ \\recovered\\ ($33.1\%$)\end{tabular}\\
            \hline
            \begin{tabular}{c}High\\BP\end{tabular}&\begin{tabular}{c}$17$ out of $183$ \\recovered\\ ($9.3\%$)\end{tabular}&\begin{tabular}{c}$3$ out of $6$ \\recovered\\ ($50.0\%$)\end{tabular}\\
            \hline
            Combined data&\begin{tabular}{c}$392$ out of $588$ \\recovered\\ ($66.7\%$)\end{tabular}&\begin{tabular}{c}$162$ out of $487$ \\recovered\\ ($33.3\%$)\end{tabular}\\
            \hline
            \end{tabular}
            
            \label{tb3} 
\end{table}

% age ?    male
% age 20-70 male female 100 samples

Note that the data in Table \ref{tb3} is observational data. The experimental data is not available yet. However, the set $\{C\}$ satisfied the back-door criterion for both $(A,Z)$ and $(A,R)$ \cite{pearl1995causal}. By Pearl's adjustment formula, the experimental data needed are: $P(r_a|c)=P(r|a,c)=0.6667$, $P(r_{a'}|c)=P(r|a',c)=0.3326$, $P(z_a|c)=P(z|a,c)=0.6888$, and $P(z'_{a'}|c)=P(z'|a',c)=0.0123$.

First, we apply Li-Pearl's Theorem to the combined data in Table \ref{tb3} and the above experimental data, the bounds of the benefit function are $[-0.3320, 0.3333]$ (see the appendix for details), and the midpoint is $0.0007$. It suggests that the drug should apply to the 30 years old male because the difference between the number of effective patients and the number of ineffective patients per 30 years old male is positive. Or someone may say that it is hard to decide because the bounded area is roughly half positive and half negative.

Second, we apply the proposed Theorem \ref{thm2} to the entire data in Table \ref{tb3} and the above experimental data, the bounds of the benefit function are $[-0.3320, -0.0054]$ (see the appendix for details), and the midpoint is $-0.1687$. The upper bound dropped significantly from $0.3333$ to $-0.0054$. It suggests that the drug should not apply to the 30 years old male, because the difference between the number of effective patients and the number of ineffective patients per 30 years old male is negative. Most importantly, the entire bounded area is negative so that the decision is convincing.

\section{Simulated Results}
\label{simres}
In this section, we will show how much in general the bounds of the benefit function are improved by Theorems \ref{thm1}, \ref{thm2}, and \ref{thm3} in three simple causal diagrams.

For each theorem, we randomly generated $100000$ sample distributions (observational data and experimental data) compatible with the causal diagram (see the appendix for the generating algorithm). Each sample distribution represents a different instantiate of the population-specific characteristics $C$ in the model. The generating algorithm ensures that the experimental data and observational data satisfy the general relation (i.e., $P(x,y|c)\le P(y|do(x),c) \le 1 - P(x, y'|c)$) \cite{tian2000probabilities}. We set the benefit vector $(\beta,\gamma,\theta,\delta)$ to be the most common $(1,-1,-1,-1)$ to encourage compliers while avoiding always-takers, never-takers, and defiers. For the sample distribution $i$, let $[a_i,b_i]$ be the bounds that considered the covariates and the causal diagram from the proposed theorems and $[c_i,d_i]$ be the bounds that did not consider the covariates and the causal diagram from Li-Pearl's Theorem. We summarized the following criteria for each case:
\begin{itemize}
    \item Average increased lower bound : $\frac{\sum(a_i-c_i)}{100000}$;
    \item Average decreased upper bound : $\frac{\sum(d_i-b_i)}{100000}$;
    \item Average gap that did not consider the covariates and the causal diagram : $\frac{\sum(d_i-c_i)}{100000}$;
    \item Average gap that considered the covariates and the causal diagram : $\frac{\sum(b_i-a_i)}{100000}$;
    \item Number of sample distributions in which the decision was flipped : $\sum e_i$ where, $e_i=1$ if $(a_i+b_i)\times (c_i+d_i) < 0$ and $e_i=0$ otherwise;
    \item Number of sample distributions in which the bounds that considered the covariates and the causal diagram from proposed Theorems were narrower : $\sum f_i$ where, $f_i=1$ if $(a_i > c_i)~or~(b_i < d_i)$ and $f_i=0$ otherwise.
\end{itemize}

\subsection{Non-descendant Covariates}
\begin{figure}[h]
\centering
\begin{tikzpicture}[->,>=stealth',node distance=2cm,
  thick,main node/.style={circle,fill,inner sep=1.5pt}]
%  \node[main node] (4) [label=above:{$U$}]{};
  \node[main node] (0) [below =1cm of 4,label=above:{$Z$}]{};
  \node[main node] (1) [below =1cm of 0,label=below:{$C$}]{};
  \node[main node] (2) [below left =1cm of 1,label=left:$X$]{};
  \node[main node] (3) [below right =1cm of 1,label=right:$Y$] {};
  \path[every node/.style={font=\sffamily\small}]
    % (0) edge node {} (2)
    % (0) edge node {} (3)
     (2) edge node {} (3);
    % (1) edge node {} (2)
    % (1) edge node {} (3);
%   \draw [dashed, -] (4) -- (2);
%   \draw [dashed, -] (4) -- (3);
   \draw [dashed, ->] (0) -- (2);
   \draw [dashed, ->] (0) -- (3);
   \draw [dashed, -] (0) -- (1);
%   \draw [dashed, -] (2) -- (3);
   \draw [dashed, ->] (1) -- (2);
   \draw [dashed, ->] (1) -- (3);
%   \draw [dashed, -] (1) -- (3);
\end{tikzpicture}
\caption{Causal diagram such that $C\cup Z$ is not a descendant of $X$.}
\label{causalg5}
\end{figure}
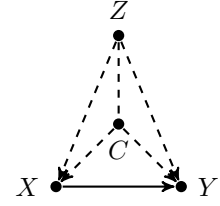

In the case of non-descendant covariates compatible with Theorem \ref{thm1}. We randomly generated $100000$ sample distributions compatible with the causal diagram in Figure \ref{causalg5}.

The results between proposed Theorem \ref{thm1} and Li-Pearl's Theorem are summarized in Table \ref{simres61}. We can see that the average gap that did not consider the covariates and the causal diagram by Li-Pearl's Theorem is $0.4342$, while the average gap that considered the covariates and the causal diagram by Theorem \ref{thm1} is $0.3352$, and both the lower bound and upper bound are improved by roughly $0.05$. The decisions flipped (i.e., the results of Li-Pearl's Theorem suggest gain profit, while the results of Theorem \ref{thm1} suggest losing profit, or the reverse) is $920/100000\approx 1\%$ of the samples, which means that at least $1\%$ of the applications would have the wrong decision if we do not consider the covariates. The bounds that considered the covariates and the causal diagram are narrower in $93688/100000\approx 93.7\%$ of the samples. Therefore, if a set of $Z$ is available that satisfies Theorem \ref{thm1}, the bounds of the benefit function by the proposed theorem are more useful as the gap is narrower. 

\begin{table}[h]
\centering
\caption{Simulation results of $100000$ sample distributions compatible with the causal diagram in Figure \ref{causalg5}.}
\begin{tabular}{|c|c|c|}
\hline
\begin{tabular}{c}Average\\increased\\lower\\bound\end{tabular}&\begin{tabular}{c}Average\\decreased\\upper\\bound\end{tabular}&\begin{tabular}{c}Average\\gap by\\Li-Pearl's\\ Theorem\end{tabular}\\
\hline
$0.0494$&$0.0496$&$0.4342$\\
\hline
\begin{tabular}{c}Average\\gap by\\Theorem \ref{thm1}\end{tabular}&\begin{tabular}{c}Decision\\flipped\end{tabular}&\begin{tabular}{c}Bounds\\narrower\end{tabular}\\
\hline
$0.3352$&$920$&$93688$\\
\hline
\end{tabular}
\vspace{5pt}
\label{simres61}
\end{table}

\begin{figure}
\centering
\includegraphics[width=0.474\textwidth]{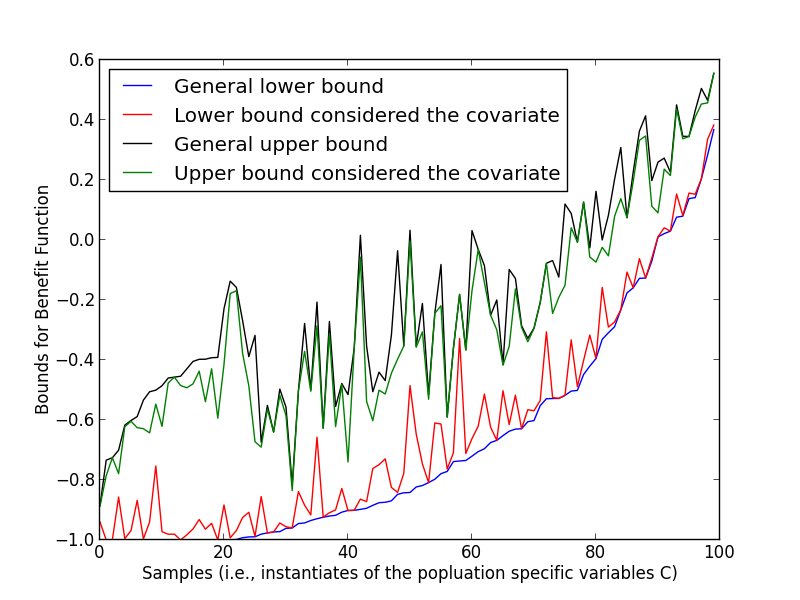}
\caption{Bounds of the benefit function for $100$ samples compatible with the causal diagram of Figure \ref{causalg5}, where the general bounds are obtained from Li-Pearl's Theorem and the bounds that considered the non-descendant covariate and the causal diagram are obtained from Theorem \ref{thm1}.}
\label{res61}
\end{figure}

We then randomly picked $100$ of $100000$ sample distributions to draw the graph of bounds that considered and did not consider the covariates and the causal diagram (To have a better vision, we sorted the sample distributions by the general lower bound that did not considered the covariates and the causal diagram). The results are shown in Figure \ref{res61}.
We can see that the bounds of the benefit function are improved in most of the samples with the causal diagram. 

\subsection{Partial Mediators}

In the case of partial mediators compatible with Theorem \ref{thm2}. We randomly generated $100000$ sample distributions that are compatible with the causal diagram in Figure \ref{causalg2}.

The results between the proposed Theorem \ref{thm2} and Li-Pearl's Theorem are summarized in Table \ref{simres62}. First, the average increased lower bound is $0$ because the lower bound in Theorem \ref{thm2} is exactly the lower bound in Li-Pearl's Theorem. The partial mediator cannot improve the lower bound. The average gap is also close between Li-Pearl's Theorem and proposed Theorem \ref{thm2} because the bounds of only $12724/100000\approx 12.7\%$ of samples are narrowed by the proposed Theorem \ref{thm2}. $12.7\%$ is an acceptable number if the costs for considering the partial mediators are acceptable. The actual improvement among the narrowed samples is impressive. We then randomly generated $100000$ samples that the bounds are indeed narrowed by the proposed Theorem \ref{thm2} (same generating algorithm, but we keep generating until we have $100000$ narrowed samples). The results of the comparison between the proposed Theorem \ref{thm2} and Li-Pearl's Theorem are summarized in Table \ref{simres63}. We can see that the average gap that did not consider the partial mediator and the causal diagram is $0.5531$, while the average gap that considered the partial mediator and the causal diagram by Theorem \ref{thm2} is $0.4768$, and the upper bound is improved by roughly $0.0764$. Therefore, if a set of $Z$ is available that satisfies Theorem \ref{thm2} and the costs permitted, we should always consider the partial mediators and using Theorem \ref{thm2}.

\begin{table}[h]
\centering
\caption{Simulation results of $100000$ sample distributions compatible with the causal diagram in Figure \ref{causalg2}.}
\begin{tabular}{|c|c|c|}
\hline
\begin{tabular}{c}Average\\increased\\lower\\bound\end{tabular}&\begin{tabular}{c}Average\\decreased\\upper\\bound\end{tabular}&\begin{tabular}{c}Average\\gap by\\Li-Pearl's\\ Theorem\end{tabular}\\
\hline
$0$&$0.00985$&$0.4564$\\
\hline
\begin{tabular}{c}Average\\gap by\\Theorem \ref{thm2}\end{tabular}&\begin{tabular}{c}Decision\\flipped\end{tabular}&\begin{tabular}{c}Bounds\\narrower\end{tabular}\\
\hline
$0.4465$&$139$&$12724$\\
\hline
\end{tabular}
\vspace{5pt}
\label{simres62}
\end{table}

\begin{table}[h]
\centering
\caption{Simulation results of $100000$ narrowed sample distributions compatible with the causal diagram in Figure \ref{causalg2}.}
\begin{tabular}{|c|c|c|}
\hline
\begin{tabular}{c}Average\\increased\\lower\\bound\end{tabular}&\begin{tabular}{c}Average\\decreased\\upper\\bound\end{tabular}&\begin{tabular}{c}Average\\gap by\\Li-Pearl's\\ Theorem\end{tabular}\\
\hline
$0$&$0.0764$&$0.5531$\\
\hline
\begin{tabular}{c}Average\\gap by\\Theorem \ref{thm2}\end{tabular}&\begin{tabular}{c}Decision\\flipped\end{tabular}&\\
\hline
$0.4768$&$1033$&\\
\hline
\end{tabular}
\vspace{5pt}
\label{simres63}
\end{table}

\begin{figure}
\centering
\includegraphics[width=0.474\textwidth]{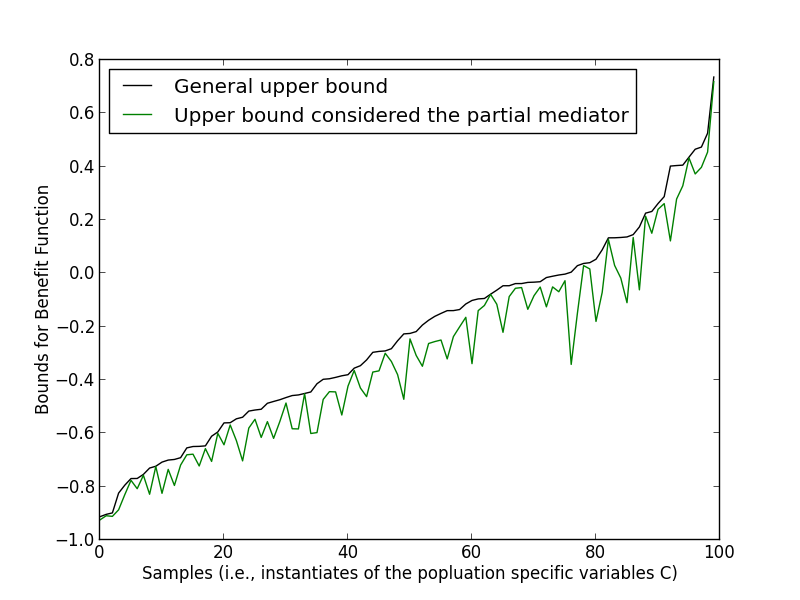}
\caption{Upper bound of the benefit function for $100$ narrowed samples compatible with the causal diagram of Figure \ref{causalg2}, where the general upper bounds are obtained from Li-Pearl's Theorem and the upper bounds that considered the partial mediator and the causal diagram are obtained from Theorem \ref{thm2}.}
\label{res62}
\end{figure}

We then randomly picked $100$ of $100000$ narrowed sample distributions to draw the graph of bounds that considered and did not considered the partial mediator and the causal diagram (To have a better vision, we sorted the sample distributions by the general upper bound that did not consider the partial mediator and the causal diagram). The results are shown in Figure \ref{res62}.
We can see that the upper bounds of the benefit function are improved significantly among these narrowed cases.

\subsection{Pure Mediators}
In the case of pure mediators compatible with Theorem \ref{thm3}. We randomly generated $100000$ sample distributions compatible with the causal diagram in Figure \ref{causalg3}.

The results between the proposed Theorem \ref{thm3} and Li-Pearl's Theorem are summarized in Table \ref{simres64}. We can see that the average gap that did not consider the pure mediator and the causal diagram by Li-Pearl's Theorem is $0.5195$, while the average gap that considered the pure mediator and the causal diagram by Theorem \ref{thm3} is $0.3324$, and the upper bound is improved by roughly $0.187$. The lower bound is not improved, because the lower bound in Theorem \ref{thm3} is exactly the same as in Li-Pearl's Theorem. The decisions flipped (i.e., the results of Li-Pearl's Theorem suggest gain profit, while the results of Theorem \ref{thm3} suggest losing profit, or the reverse) is $459/100000\approx 0.46\%$ of the samples, which means that at least $0.46\%$ of the applications would have the wrong decision if we do not consider the pure mediators. The bounds that considered the pure mediator and the causal diagram are narrower in $99996/100000\approx 99.9\%$ of the samples. Therefore, if a set of $Z$ is available that satisfies Theorem \ref{thm3}, the bounds of the benefit function by the proposed theorem is more useful as the gap is narrower. 

\begin{table}[h]
\centering
\caption{Simulation results of $100000$ sample distributions compatible with the causal diagram in Figure \ref{causalg3}.}
\begin{tabular}{|c|c|c|}
\hline
\begin{tabular}{c}Average\\increased\\lower\\bound\end{tabular}&\begin{tabular}{c}Average\\decreased\\upper\\bound\end{tabular}&\begin{tabular}{c}Average\\gap by\\Li-Pearl's\\ Theorem\end{tabular}\\
\hline
$0$&$0.1870$&$0.5195$\\
\hline
\begin{tabular}{c}Average\\gap by\\Theorem \ref{thm3}\end{tabular}&\begin{tabular}{c}Decision\\flipped\end{tabular}&\begin{tabular}{c}Bounds\\narrower\end{tabular}\\
\hline
$0.3324$&$459$&$99996$\\
\hline
\end{tabular}
\vspace{5pt}
\label{simres64}
\end{table}

\begin{figure}
\centering
\includegraphics[width=0.474\textwidth]{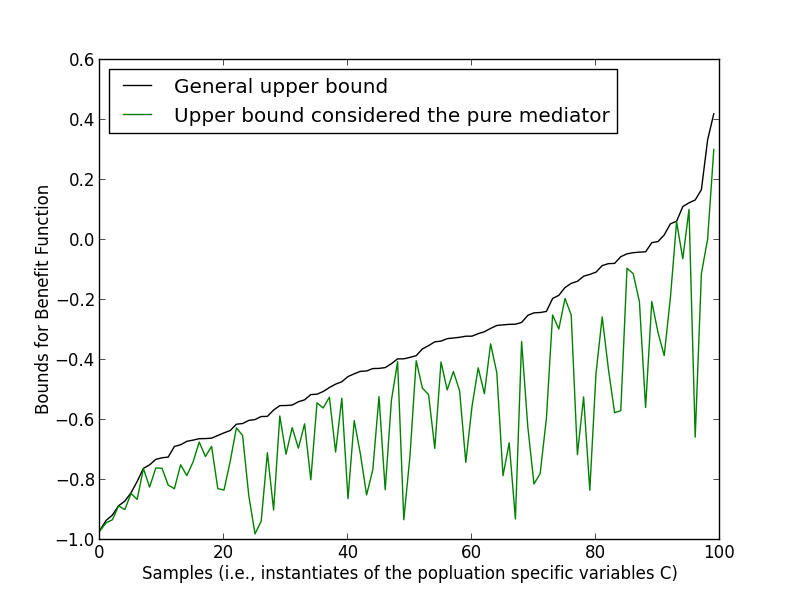}
\caption{Bounds of the benefit function for $100$ samples compatible with the causal diagram of Figure \ref{causalg3}, where the general bounds are obtained from Li-Pearl's Theorem and the bounds that considered the pure mediator and the causal diagram are obtained from Theorem \ref{thm3}.}
\label{res63}
\end{figure}

We then randomly picked $100$ of $100000$ sample distributions to draw the graph of bounds that considered and did not consider the pure mediator and the causal diagram (To have a better vision, we sorted the sample distributions by the general upper bound that did not consider the pure mediator and the causal diagram). The results are shown in Figure \ref{res63}.
We can see that the bounds of the benefit function are improved in almost all the samples with the causal diagram. 

\section{Discussion}
In this section, we will discuss one more requirement of covariates $Z$ in Theorem \ref{thm1}. Note that in the motivating example in the introduction section, the discount should apply to the entire company A's employees; the carwash company can only decide to offer the discount to the entire company A or not to the entire company A. The carwash company cannot offer the discount to a specific age group in company A. Otherwise, if the carwash company can offer the discount to a specific age group, the covariates $Z$ should be considered as the population-specific characteristics and combined into $C$, and apply Li-Pearl's Theorem separately to each population-specific group. This requirement is common; for example, an election speech cannot offer to only a specific group of people in a region, and an auto show cannot offer to only a specific group of customers in a region. This requirement does not apply to Theorems \ref{thm2} and \ref{thm3} because the mediators happen after the treatment.
% when should consider the covariates by covaraites attribution

\section{Conclusion}
We demonstrated how bounds of the benefit function in the unit selection problem could be narrowed if covariates information and their associated causal structures are available. We derived three theorems to narrow the bounds of the benefit function in three common graphical conditions. We illustrated that if costs are permitted, and there are covariates and causal structures available, the proposed theorems should always be applied, as narrower bounds are helping to make accurate decisions. Examples and simulation results are provided to support the proposed theorems.

\bibliography{aaai22}

\begin{thebibliography}{13}
\providecommand{\natexlab}[1]{#1}

\bibitem[{Berson, Smith, and Thearling(1999)}]{berson1999building}
Berson, A.; Smith, S.; and Thearling, K. 1999.
\newblock \emph{Building data mining applications for CRM}.
\newblock McGraw-Hill Professional.

\bibitem[{Bottou et~al.(2013)Bottou, Peters, Qui{\~n}onero-Candela, Charles,
  Chickering, Portugaly, Ray, Simard, and Snelson}]{bottou2013counterfactual}
Bottou, L.; Peters, J.; Qui{\~n}onero-Candela, J.; Charles, D.~X.; Chickering,
  D.~M.; Portugaly, E.; Ray, D.; Simard, P.; and Snelson, E. 2013.
\newblock Counterfactual reasoning and learning systems: The example of
  computational advertising.
\newblock \emph{The Journal of Machine Learning Research}, 14(1): 3207--3260.

\bibitem[{Dawid, Musio, and Murtas(2017)}]{dawid2017}
Dawid, P.; Musio, M.; and Murtas, R. 2017.
\newblock The Probability of Causation.
\newblock \emph{Law, Probability and Risk}, (16): 163--179.

\bibitem[{Hung, Yen, and Wang(2006)}]{hung2006applying}
Hung, S.-Y.; Yen, D.~C.; and Wang, H.-Y. 2006.
\newblock Applying data mining to telecom churn management.
\newblock \emph{Expert Systems with Applications}, 31(3): 515--524.

\bibitem[{Lejeune(2001)}]{lejeune2001measuring}
Lejeune, M.~A. 2001.
\newblock Measuring the impact of data mining on churn management.
\newblock \emph{Internet Research}, 11(5): 375--387.

\bibitem[{Li and Pearl(2019)}]{li2019unit}
Li, A.; and Pearl, J. 2019.
\newblock Unit selection based on counterfactual logic.
\newblock In \emph{Proceedings of the 28th International Joint Conference on
  Artificial Intelligence}, 1793--1799. AAAI Press.

\bibitem[{Li et~al.(2014)Li, Chen, Kleban, and Gupta}]{li2014counterfactual}
Li, L.; Chen, S.; Kleban, J.; and Gupta, A. 2014.
\newblock Counterfactual estimation and optimization of click metrics for
  search engines.
\newblock \emph{arXiv preprint arXiv:1403.1891}.

\bibitem[{Mueller, Li, and Pearl(2021)}]{mueller2021causes}
Mueller, S.; Li, A.; and Pearl, J. 2021.
\newblock Causes of Effects: Learning individual responses from population
  data.
\newblock \emph{arXiv preprint arXiv:2104.13730}.

\bibitem[{Pearl(1995)}]{pearl1995causal}
Pearl, J. 1995.
\newblock Causal diagrams for empirical research.
\newblock \emph{Biometrika}, 82(4): 669--688.

\bibitem[{Sun et~al.(2015)Sun, Wang, Yin, Yang, and Chang}]{sun2015causal}
Sun, W.; Wang, P.; Yin, D.; Yang, J.; and Chang, Y. 2015.
\newblock Causal inference via sparse additive models with application to
  online advertising.
\newblock In \emph{AAAI}, 297--303.

\bibitem[{Tian and Pearl(2000)}]{tian2000probabilities}
Tian, J.; and Pearl, J. 2000.
\newblock Probabilities of causation: Bounds and identification.
\newblock \emph{Annals of Mathematics and Artificial Intelligence}, 28(1-4):
  287--313.

\bibitem[{Tsai and Lu(2009)}]{tsai2009customer}
Tsai, C.-F.; and Lu, Y.-H. 2009.
\newblock Customer churn prediction by hybrid neural networks.
\newblock \emph{Expert Systems with Applications}, 36(10): 12547--12553.

\bibitem[{Yan et~al.(2009)Yan, Liu, Wang, Zhang, Jiang, and Chen}]{yan2009much}
Yan, J.; Liu, N.; Wang, G.; Zhang, W.; Jiang, Y.; and Chen, Z. 2009.
\newblock How much can behavioral targeting help online advertising?
\newblock In \emph{Proceedings of the 18th international conference on World
  Wide Web}, 261--270. ACM.

\end{thebibliography}
\clearpage
\newpage
\appendix
\section{Appendix}
\subsection{Proof of Theorems}
First, we have the following Lemmas \ref{lm6} and \ref{lm7} from \cite{li2019unit}.

\begin{lemma}
The c-specific PNS $P(y_x,y'_{x'}|c)$ is bounded as follows:
\begin{eqnarray}
&&max\left\{
\begin{array}{c}
0,\\
P(y_x|c)-P(y_{x'}|c),\\
P(y|c)-P(y_{x'}|c),\\
P(y_x|c)-P(y|c)
\end{array}
\right\}\le \text{c-PNS},\nonumber\\
%\label{inequ11}\\
&&min\left\{
\begin{array}{c}
P(y_x|c),\\
P(y'_{x'}|c),\\
P(y,x|c)+P(y',x'|c),\\
P(y_x|c)-P(y_{x'}|c)+\\
+P(y,x'|c)+P(y',x|c)
\end{array}
\right\}\ge \text{c-PNS}.\nonumber
%\label{inequ22}
\end{eqnarray}
\label{lm6}
\end{lemma}

\begin{lemma}
\begin{eqnarray}
&&P(y_x,y'_{x'}|c)-P(y'_x,y_{x'}|c) \nonumber \\
&=&P(y_x|c)-P(y_{x'}|c).\nonumber
%\label{hmmm}
\end{eqnarray}
% \begin{proof}
% \begin{eqnarray*}
% &&P(y_x,y'_{x'}|c)-P(y_{x'},y'_x|c)\\
% &=&P(y_x,y'_{x'},x|c)+P(y_x,y'_{x'},x'|c)-P(y_{x'},y'_{x},x|c)-P(y_{x'},y'_{x},x'|c) \\
% &=&P(y,y'_{x'},x|c)+P(y_x,y',x'|c)-P(y_{x'},y',x|c)-P(y,y'_{x},x'|c) \\
% &=&P(y,y'_{x'},x|c)-P(y_{x'},y',x|c)+P(y_x,y',x'|c)-P(y,y'_{x},x'|c) \\
% &=&P(x,y|c)-P(y,y_{x'},x|c)-P(y_{x'},y',x|c)+P(y_x,y',x'|c)+P(y,y_{x},x'|c)-P(x',y|c)\\
% &=&P(x,y|c)-P(y_{x'},x|c)+P(y_x,x'|c)-P(x',y|c)\\
% &=&P(x,y|c)-P(y_{x'}|c)+P(y_{x'},x'|c)+P(y_x|c)-P(y_x,x|c)-P(x',y|c)\\
% &=&P(x,y|c)-P(y_{x'}|c)+P(y,x'|c)+P(y_x|c)-P(y,x|c)-P(x',y|c)\\
% &=&P(y_x|c)-P(y_{x'}|c).
% \end{eqnarray*}
% \end{proof}
\label{lm7}
\end{lemma}

\begin{lemma}
Given a causal diagram $G$ and distribution compatible with $G$, let $Z\cup C$ be a set of variables that does not contain any descendant of $X$ in $G$, then c-specific PNS $P(y_x,y'_{x'}|c)$ is bounded as follows:
\begin{eqnarray}
&&\sum_z\max\left\{
\begin{array}{c}
0,\\
P(y_x|z,c)-P(y_{x'}|z,c),\\
P(y|z,c)-P(y_{x'}|z,c),\\
P(y_x|z,c)-P(y|z,c)
\end{array}
\right\}\nonumber\\
&&\times P(z|c) \le c \text{-PNS,}
\label{ineqch71}\\
&&\sum_z\min\left\{
\begin{array}{c}
P(y_x|z,c),\\
P(y'_{x'}|z,c),\\
P(y,x|z,c)+P(y',x'|z,c),\\
P(y_x|z,c)-P(y_{x'}|z,c)+\\
+P(y,x'|z,c)+P(y',x|z,c)
\end{array}
\right\}\nonumber\\
&&\times P(z|c)\ge c \text{-PNS.}
\label{ineqch72}
\end{eqnarray}
\label{lmch71}
\begin{proof}
\begin{eqnarray}
c\text{-PNS} & = & P(y_x,y'_{x'}|c)\nonumber \\
&=& \sum_z P(y_x,y'_{x'}|z,c)\times P(z|c).
\label{thma65p1}
\end{eqnarray}
From Lemma \ref{lm6}, replace $c$ with $(z,c)$, we have the following:
\begin{eqnarray}
&&\max\left\{
\begin{array}{c}
0,\\
P(y_x|z,c)-P(y_{x'}|z,c),\\
P(y|z,c)-P(y_{x'}|z,c),\\
P(y_x|z,c)-P(y|z,c)
\end{array}
\right\}\nonumber\\
&&\le P(y_x,y'_{x'}|z,c),
\label{inequa6111}\\
&&\min\left\{
\begin{array}{c}
P(y_x|z,c),\\
P(y'_{x'}|z,c),\\
P(y,x|z,c)+P(y',x'|z,c),\\
P(y_x|z,c)-P(y_{x'}|z,c)+\\
+P(y,x'|z,c)+P(y',x|z,c)
\end{array}
\right\}\nonumber\\
&&\ge P(y_x,y'_{x'}|z,c).
\label{inequa6222}
\end{eqnarray}
Substituting Equations \ref{inequa6111} and \ref{inequa6222} into Equation \ref{thma65p1}, Lemma \ref{lmch71} holds.\\
Note that since we have,\\
\begin{eqnarray}
&&\sum_z \max\{0,\nonumber\\
&&P(y_x|z,c)-P(y_{x'}|z,c),\nonumber\\
&&P(y|z,c)-P(y_{x'}|z,c),\nonumber\\
&&P(y_x|z,c)-P(y|z,c)\}\times P(z|c) \nonumber\\
&\ge& \sum_z 0\times P(z|c) \nonumber \\
&=& 0, \nonumber \\
\nonumber \\
&&\sum_z \max\{0,\nonumber\\
&&P(y_x|z,c)-P(y_{x'}|z,c),\nonumber \\
&&P(y|z,c)-P(y_{x'}|z,c),\nonumber\\
&&P(y_x|z,c)-P(y|z,c)\}\times P(z|c) \nonumber\\
&\ge& \sum_z [P(y_x|z,c)-P(y_{x'}|z,c)]\times P(z|c) \nonumber \\
&=& P(y_x|c)-P(y_{x'}|c), \nonumber \\
\nonumber \\
&&\sum_z \max\{0,\nonumber\\
&&P(y_x|z,c)-P(y_{x'}|z,c),\nonumber \\
&&P(y|z,c)-P(y_{x'}|z,c),\nonumber\\
&&P(y_x|z,c)-P(y|z,c)\}\times P(z|c) \nonumber\\
&\ge& \sum_z [P(y|z,c)-P(y_{x'}|z,c)]\times P(z|c) \nonumber \\
&=& P(y|c)-P(y_{x'}|c), \nonumber \\
\nonumber \\
&&\sum_z \max\{0,\nonumber\\
&&P(y_x|z,c)-P(y_{x'}|z,c),\nonumber \\
&&P(y|z,c)-P(y_{x'}|z,c),\nonumber\\
&&P(y_x|z,c)-P(y|z,c)\}\times P(z|c)\nonumber\\
&\ge& \sum_z [P(y_x|z,c)-P(y|z,c)]\times P(z|c)\nonumber\\
&=& P(y_x|c)-P(y|c),\nonumber
\end{eqnarray}
then the lower bound in Lemma \ref{lmch71} is guaranteed to be no worse than the lower bound in Lemma \ref{lm6}. Similarly, the upper bound in Lemma \ref{lmch71} is guaranteed to be no worse than the upper bound in Lemma \ref{lm6}. Also note that, since $Z\cup C$ does not contain a descendant of $X$, the term $P(y_x|z,c)$ refers to experimental data under population $z,c$.
\end{proof}
\end{lemma}

\begin{lemma}
\begin{eqnarray}
f(c)&=&\beta P(y_x,y'_{x'}|c)+\gamma P(y_x,y_{x'}|c)+\nonumber\\
&&+ \theta P(y'_x,y'_{x'}|c) + \delta P(y_{x'},y'_{x}|c)\nonumber\\  
&=&W +\sigma P(y_x,y'_{x'}|c).
\label{eqb0}
\end{eqnarray}
where,
\begin{eqnarray*}
&&W=(\gamma -\delta)P(y_x|c)+\delta P(y_{x'}|c)+\theta P(y'_{x'}|c),\\
&&\sigma=\beta -\gamma-\theta +\delta.
\end{eqnarray*}
\begin{proof}
\begin{eqnarray}
&&f(c)\nonumber\\
&=&\beta P(y_x,y'_{x'}|c)+\gamma P(y_{x},y_{x'}|c)+\nonumber\\
&&+\theta P(y'_x,y'_{x'}|c)+\delta P(y'_{x},y_{x'}|c)\nonumber \\
&=&\beta P(y_x,y'_{x'}|c)+\gamma [P(y_x|c)-P(y_{x},y'_{x'}|c)]+\nonumber\\
&&+\theta [P(y'_{x'})-P(y_x,y'_{x'}|c)]+\delta P(y'_{x},y_{x'}|c)\nonumber \\
&=&\gamma P(y_x|c)+\theta P(y'_{x'}|c)+\nonumber\\
&&+(\beta -\gamma -\theta) P(y_x,y'_{x'}|c)+\delta P(y'_{x},y_{x'}|c).
\label{eqb1}
\end{eqnarray}
By Lemma \ref{lm7}, we have,
\begin{eqnarray}
P(y'_x,y_{x'}|c)=P(y_x,y'_{x'}|c)-P(y_x|c)+P(y_{x'}|c).
\label{eqb2}
\end{eqnarray}
Substituting Equation \ref{eqb2} into Equation \ref{eqb1}, we have,
\begin{eqnarray}
&&f(c)\nonumber\\
&=&\gamma P(y_x|c)+\theta P(y'_{x'}|c)+\nonumber\\
&&+(\beta -\gamma -\theta) P(y_x,y'_{x'}|c)+\delta P(y'_{x},y_{x'}|c)\nonumber\\
&=&\gamma P(y_x|c)+\theta P(y'_{x'}|c)+\nonumber\\
&&+(\beta -\gamma -\theta) P(y_x,y'_{x'}|c)+\nonumber\\
&&+\delta [P(y_x,y'_{x'}|c)-P(y_x|c)+P(y_{x'}|c)]\nonumber\\
&=&(\gamma -\delta)P(y_x|c)+\delta P(y_{x'}|c)+\theta P(y'_{x'}|c) +\nonumber\\
&&+(\beta -\gamma-\theta +\delta) P(y_x,y'_{x'}|c).\nonumber
%\label{eqb3}
\end{eqnarray}
\end{proof}
\label{lmch72}
\end{lemma}

\begin{reptheorem}{thm1}
Given a causal diagram $G$ and distribution compatible with $G$, let $Z\cup C$ be a set of variables that does not contain any descendant of $X$ in $G$, then the benefit function $f(c)=\beta P(y_x,y'_{x'}|c)+\gamma P(y_x,y_{x'}|c)+ \theta P(y'_x,y'_{x'}|c) + \delta P(y_{x'},y'_{x}|c)$ is bounded as follows:
\begin{eqnarray*}
W+\sigma U\le f \le W+\sigma L\text{~~~~~~~~if }\sigma < 0,\\
W+\sigma L\le f \le W+\sigma U\text{~~~~~~~~if }\sigma > 0,
\end{eqnarray*}
where $\sigma, W,L,U$ are given by,
\begin{eqnarray*}
&&\sigma = \beta - \gamma - \theta + \delta,\\
&&W=(\gamma -\delta)P(y_x|c)+\delta P(y_{x'}|c)+\theta P(y'_{x'}|c),\\
&&L=\sum_z\max\left\{
\begin{array}{c}
0,\\
P(y_x|z,c)-P(y_{x'}|z,c),\\
P(y|z,c)-P(y_{x'}|z,c),\\
P(y_x|z,c)-P(y|z,c)
\end{array}
\right\}\nonumber\\
&&\times P(z|c),\\
&&U=\sum_z\min\left\{
\begin{array}{c}
P(y_x|z,c),\\
P(y'_{x'}|z,c),\\
P(y,x|z,c)+P(y',x'|z,c),\\
P(y_x|z,c)-P(y_{x'}|z,c)+\\
+P(y,x'|z,c)+P(y',x|z,c)
\end{array}
\right\}\nonumber\\
&&\times P(z|c).
\end{eqnarray*}
\begin{proof}
By Lemmas \ref{lmch71} and \ref{lmch72},\\
substituting Equations \ref{ineqch71} and \ref{ineqch72} into Equation \ref{eqb0}, Theorem \ref{thm1} holds.

Note that, if we substituting Lemma \ref{lm6} into Lemma \ref{lmch72}, we have the same results as in Li-Pearl's Theorem. We showed that in Lemma \ref{lmch71} that the bounds in Lemma \ref{lmch71} is guaranteed to be no worse than the bounds in Lemma \ref{lm6}, therefore, the bounds in Theorem \ref{thm1} is guaranteed to be no worse than the bounds in Li-Pearl's Theorem.
\end{proof}
\end{reptheorem}

\begin{lemma}
Given a causal diagram $G$ and distribution compatible with $G$, let $Z\cup C$ be a set of variables such that $\forall x,x' \in X : x \ne x', (Y_x \independent X\cup Z_{x'}\ |\ Z_x,C)$ in $G$, then the c-PNS $P(y_x,y'_{x'}|c)$ is bounded as follows:
\begin{eqnarray}
&&\max\left\{
\begin{array}{c}
0,\\
P(y_x|c)-P(y_{x'}|c),\\
P(y|c)-P(y_{x'}|c),\\
P(y_x|c)-P(y|c)\\
\end{array}
\right\}\le c\text{-PNS,}
\label{ineqch73}\\
&&\min\left\{
\begin{array}{c}
P(y_x|c),\\
P(y'_{x'}|c),\\
P(y,x|c)+P(y',x'|c),\\
P(y_x|c)-P(y_{x'}|c)+\\
+P(y,x'|c)+P(y',x|c),\\
\sum_z \sum_{z'} \min\{P(y|z,x,c),\\
P(y'|z',x',c)\}\\
\times \min\{P(z_x|c),P(z'_{x'}|c)\}
\end{array}
\right\}\ge c\text{-PNS.}\nonumber\\
\label{ineqch74}
\end{eqnarray}
\label{lmch73}
\begin{proof}
\begin{eqnarray}
&&c\text{-PNS} \nonumber\\
&=& P(y_x,y'_{x'}|c)\nonumber\\
&=& \Sigma_z \Sigma_{z'} P(y_x,y'_{x'},z_x,z'_{x'}|c) \nonumber\\
&=& \Sigma_z \Sigma_{z'} P(y_x,y'_{x'}|z_x,z'_{x'},c)\times P(z_x,z'_{x'}|c) \nonumber\\
&\le& \Sigma_z \Sigma_{z'} \min\{P(y_x|z_x,z'_{x'},c), P(y'_{x'}|z_x,z'_{x'},c)\}\nonumber\\
&&\times \min\{P(z_x|c),P(z'_{x'}|c)\}\nonumber\\
&=& \Sigma_z \Sigma_{z'} \min\{P(y_x|z_x,c), P(y'_{x'}|z'_{x'},c)\}\nonumber\\
&&\times\min\{P(z_x|c),P(z'_{x'}|c)\} \label{part_mediator_zx}\\
&=& \Sigma_z \Sigma_{z'} \min\{P(y|z_x,x,c), P(y'|z'_{x'},x',c)\}\nonumber\\
&&\times\min\{P(z_x|c),P(z'_{x'}|c)\} \label{part_mediator_x}\\
&=& \Sigma_z \Sigma_{z'} \min\{P(y|z,x,c), P(y'|z',x',c)\}\nonumber\\
&&\times\min\{P(z_x|c),P(z'_{x'}|c)\}.\nonumber
\end{eqnarray}
Combined with the bounds in Lemma \ref{lm6}, Lemma \ref{lmch73} holds. Note that Equation \ref{part_mediator_zx} is due to $Y_x \independent Z_{x'}\ |\ Z_x,C$ and $Y_{x'} \independent Z_x\ |\ Z_{x'},C$. Equation \ref{part_mediator_x} is due to $\forall x \in X, Y_x \independent\ X\ | Z_x,C$.
\end{proof}
\end{lemma}

\begin{reptheorem}{thm2}
Given a causal diagram $G$ and distribution compatible with $G$, let $Z$ be a set of variables such that $\forall x,x' \in X : x \ne x', (Y_x \independent X\cup Z_{x'}\ |\ Z_x, C)$ in $G$, and $C$ does not contain any descendant of $X$ in $G$, then the benefit function $f(c)=\beta P(y_x,y'_{x'}|c)+\gamma P(y_x,y_{x'}|c)+ \theta P(y'_x,y'_{x'}|c) + \delta P(y_{x'},y'_{x}|c)$ is bounded as follows:
\begin{eqnarray*}
W+\sigma U\le f \le W+\sigma L\text{~~~~~~~~if }\sigma < 0,\\
W+\sigma L\le f \le W+\sigma U\text{~~~~~~~~if }\sigma > 0,
\end{eqnarray*}
where $\sigma, W,L,U$ are given by,
\begin{eqnarray*}
&&\sigma = \beta - \gamma - \theta + \delta,\\
&&W=(\gamma -\delta)P(y_x|c)+\delta P(y_{x'}|c)+\theta P(y'_{x'}|c),\\
&&L=\max\left\{
\begin{array}{c}
0,\\
P(y_x|c)-P(y_{x'}|c),\\
P(y|c)-P(y_{x'}|c),\\
P(y_x|c)-P(y|c)\\
\end{array}
\right\},\\
&&U=\min\left\{
\begin{array}{c}
P(y_x|c),\\
P(y'_{x'}|c),\\
P(y,x|c)+P(y',x'|c),\\
P(y_x|c)-P(y_{x'}|c)+\\
+P(y,x'|c)+P(y',x|c),\\
\sum_z \sum_{z'} \min\{P(y|z,x,c),\\
P(y'|z',x',c)\}\\
\times \min\{P(z_x|c),P(z'_{x'}|c)\}
\end{array}
\right\}.
\end{eqnarray*}
\begin{proof}
By Lemmas \ref{lmch73} and \ref{lmch72},\\
substituting Equations \ref{ineqch73} and \ref{ineqch74} into Equation \ref{eqb0}, Theorem \ref{thm2} holds.

Note that, if we substituting Lemma \ref{lm6} into Lemma \ref{lmch72}, we have the same results as in Li-Pearl's Theorem. From the proof of Lemma \ref{lmch73}, we know that the lower bound in Lemma \ref{lmch73} is the same as in Lemma \ref{lm6} and the upper bound in Lemma \ref{lmch73} is no worse than the upper bound in Lemma \ref{lm6}. Therefore, the lower bound in Theorem \ref{thm2} is the same as in Li-Pearl's Theorem, and the upper bound in Theorem \ref{thm2} is guaranteed to be no worse than the upper bound in Li-Pearl's Theorem.
\end{proof}
\end{reptheorem}

\begin{lemma}
Given a causal diagram $G$ in Figure \ref{causalgach65} and distribution that compatible with $G$, and $C$ is not a descendant of $X$, then c-PNS $P(y_x,y'_{x'}|c)$ is bounded as follow:

\begin{figure}[h]
\centering
\begin{tikzpicture}[->,>=stealth',node distance=2cm,
  thick,main node/.style={circle,fill,inner sep=1.5pt}]
  \node[main node] (0) [label=above:{$C$}]{};
  \node[main node] (1) [below =1cm of 0,label=below:{$Z$}]{};
  \node[main node] (2) [below left =1cm of 1,label=left:$X$]{};
  \node[main node] (3) [below right =1cm of 1,label=right:$Y$] {};
  \path[every node/.style={font=\sffamily\small}]
    (2) edge node {} (1)
    (1) edge node {} (3);
  \draw [dashed, -] (0) -- (2);
  \draw [dashed, -] (0) -- (3);
  \draw [dashed, -] (0) -- (1);
\end{tikzpicture}
\caption{Mediator $Z$ with no direct effects of $X$ on $Y$.}
\label{causalgach65}
\end{figure}
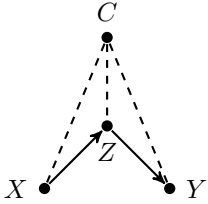
\begin{eqnarray}
&&\max\left\{
\begin{array}{c}
0,\\
P(y_x|c)-P(y_{x'}|c),\\
P(y|c)-P(y_{x'}|c),\\
P(y_x|c)-P(y|c)\\
\end{array}
\right\}\le c\text{-PNS,}
\label{ineqch75}\\
&&\min\left\{
\begin{array}{c}
P(y_x|c),\\
P(y'_{x'}|c),\\
P(y,x|c)+P(y',x'|c),\\
P(y_x|c)-P(y_{x'}|c)+\\
+P(y,x'|c)+P(y',x|c),\\
\Sigma_z \Sigma_{z'\ne z}\min\{P(y|z,c),\\
P(y'|z',c)\}\\
\times \min\{P(z|x,c),P(z'|x',c)\}
\end{array}
\right\}\ge c\text{-PNS.}\nonumber\\
\label{ineqch76}
\end{eqnarray}
\label{lmch74}
\begin{proof}
First we show that in graph $G$, if an individual is a c-complier from $X$ to $Y$, then $Z_x|c$ and $Z_{x'}|c$ must have the different values. This is because the structural equations for $Y$ and $Z$ are $f_y(z,u_y,c)$ and $f_z(x,u_z,c)$, respectively. If an individual has the same $Z_x|c$ and $Z_{x'}|c$ value, then $f_z(x,u_z,c)=f_z(x',u_z,c)$. This means $f_y(f_z(x,u_z,c),u_y,c)=f_y(f_z(x',u_z,c),u_y,c)$, i.e., $Y_x|c$ and $Y_{x'}|c$ must have the same value. Thus this individual is not a c-complier. Therefore,
\begin{eqnarray}
&&c\text{-PNS} \nonumber\\
&=&P(y_x,y'_{x'}|c) \nonumber \\
&=&\Sigma_z \Sigma_{z'\ne z}P(y_z,y'_{z'}|c)\times P(z_x,z'_{x'}|c) \nonumber \\
&\le&\Sigma_z \Sigma_{z'\ne z}\min\{P(y_z|c),P(y'_{z'}|c)\}\nonumber\\
&&\times \min\{P(z_x|c),P(z'_{x'}|c)\} \nonumber \\
&=&\Sigma_z \Sigma_{z'\ne z}\min\{P(y|z,c),P(y'|z',c)\}\nonumber\\
&&\times\min\{P(z|x,c),P(z'|x',c)\}.\nonumber
\end{eqnarray}
Combined with the bounds in Lemma \ref{lm6}, Lemma \ref{lmch74} holds.
\end{proof}
\end{lemma}

\begin{reptheorem}{thm3}
Given a causal diagram $G$ in Figure \ref{causalgach65} and distribution compatible with $G$, and $C$ does not contain any descendant of $X$, then the benefit function $f(c)=\beta P(y_x,y'_{x'}|c)+\gamma P(y_x,y_{x'}|c)+ \theta P(y'_x,y'_{x'}|c) + \delta P(y_{x'},y'_{x}|c)$ is bounded as follows:
% \begin{figure}[h]
% \centering
% \begin{tikzpicture}[->,>=stealth',node distance=2cm,
%   thick,main node/.style={circle,fill,inner sep=1.5pt}]
%   \node[main node] (0) [label=above:{$C$}]{};
%   \node[main node] (1) [below =1cm of 0,label=below:{$Z$}]{};
%   \node[main node] (2) [below left =1cm of 1,label=left:$X$]{};
%   \node[main node] (3) [below right =1cm of 1,label=right:$Y$] {};
%   \path[every node/.style={font=\sffamily\small}]
%     (2) edge node {} (1)
%     (1) edge node {} (3);
%   \draw [dashed, -] (0) -- (2);
%   \draw [dashed, -] (0) -- (3);
%   \draw [dashed, -] (0) -- (1);
% \end{tikzpicture}
% \caption{Mediator $Z$ with no direct effects.}
% \label{causalg65}
% \end{figure}
\begin{eqnarray*}
W+\sigma U\le f \le W+\sigma L\text{~~~~~~~~if }\sigma < 0,\\
W+\sigma L\le f \le W+\sigma U\text{~~~~~~~~if }\sigma > 0,
\end{eqnarray*}
where $\sigma, W,L,U$ are given by,
\begin{eqnarray*}
&&\sigma = \beta - \gamma - \theta + \delta,\\
&&W=(\gamma -\delta)P(y_x|c)+\delta P(y_{x'}|c)+\theta P(y'_{x'}|c),\\
&&L=\max\left\{
\begin{array}{c}
0,\\
P(y_x|c)-P(y_{x'}|c),\\
P(y|c)-P(y_{x'}|c),\\
P(y_x|c)-P(y|c)\\
\end{array}
\right\},\\
&&U=\min\left\{
\begin{array}{c}
P(y_x|c),\\
P(y'_{x'}|c),\\
P(y,x|c)+P(y',x'|c),\\
P(y_x|c)-P(y_{x'}|c)+\\
+P(y,x'|c)+P(y',x|c),\\
\Sigma_z \Sigma_{z'\ne z}\min\{P(y|z,c),\\
P(y'|z',c)\}\\
\times \min\{P(z|x,c),P(z'|x',c)\}
\end{array}
\right\}.
\end{eqnarray*}
\begin{proof}
By Lemmas \ref{lmch74} and \ref{lmch72},\\
substituting Equations \ref{ineqch75} and \ref{ineqch76} into Equation \ref{eqb0}, Theorem \ref{thm3} holds.

Note that, if we substituting Lemma \ref{lm6} into Lemma \ref{lmch72}, we have the same results as in Li-Pearl's Theorem. From the proof of Lemma \ref{lmch74}, we know that the lower bound in Lemma \ref{lmch74} is the same as in Lemma \ref{lm6} and the upper bound in Lemma \ref{lmch74} is no worse than the upper bound in Lemma \ref{lm6}. Therefore, the lower bound in Theorem \ref{thm3} is the same as in Li-Pearl's Theorem, and the upper bound in Theorem \ref{thm3} is guaranteed to be no worse than the upper bound in Li-Pearl's Theorem.
\end{proof}
\end{reptheorem}

\newpage
\subsection{Calculation in the Examples}
In order to clearly see the calculation steps, we list an equivalent form of Li-Pearl's Theorem as following (see the proof in the previous section for the equivalence):
\begin{theorem}
\label{thm4}
Given a causal diagram $G$ and distribution compatible with $G$, let $C$ be a set of variables that does not contain any descendant of $X$ in $G$, then the benefit function $f(c)=\beta P(y_x,y'_{x'}|c)+\gamma P(y_x,y_{x'}|c)+ \theta P(y'_x,y'_{x'}|c) + \delta P(y_{x'},y'_{x}|c)$ is bounded as follows:
\begin{eqnarray*}
W+\sigma U\le f(c) \le W+\sigma L\text{~~~~~~~~if }\sigma < 0,\\
W+\sigma L\le f(c) \le W+\sigma U\text{~~~~~~~~if }\sigma > 0,
\end{eqnarray*}
where $\sigma, W,L,U$ are given by,
\begin{eqnarray*}
&&\sigma = \beta - \gamma - \theta + \delta,\\
&&W=(\gamma -\delta)P(y_x|c)+\delta P(y_{x'}|c)+\theta P(y'_{x'}|c),\\
&&L=\max\left\{
\begin{array}{c}
0,\\
P(y_x|c)-P(y_{x'}|c),\\
P(y|c)-P(y_{x'}|c),\\
P(y_x|c)-P(y|c)\\
\end{array}
\right\},\\
&&U=\min\left\{
\begin{array}{c}
P(y_x|c),\\
P(y'_{x'}|c),\\
P(y,x|c)+P(y',x'|c),\\
P(y_x|c)-P(y_{x'}|c)+\\+P(y,x'|c)+P(y',x|c)
\end{array}
\right\}.
\end{eqnarray*}
\end{theorem}
\subsubsection{Company Selection}
First, we apply Li-Pearl's Theorem (Theorem \ref{thm4}) to the data in Tables \ref{tb1} and \ref{tb2}. The benefit vector is $(100,-60,0,-140)$.\\
We have,
\begin{eqnarray*}
\sigma &=& \beta - \gamma - \theta + \delta\\
&=&100-(-60)-0+(-140)\\
&=&20
\end{eqnarray*}
\begin{eqnarray*}
W&=&(\gamma -\delta)P(r_a|c)+\delta P(r_{a'}|c)+\theta P(r'_{a'}|c)\\
&=&(-60-(-140))\times 0.83729+0\times 0.47405+\\
&&+(-140)\times 0.52595\\
&=&-6.64980
\end{eqnarray*}
\begin{eqnarray*}
L&=&\max\left\{
\begin{array}{c}
0,\\
P(r_a|c)-P(r_{a'}|c),\\
P(r|c)-P(r_{a'}|c),\\
P(r_a|c)-P(r|c)\\
\end{array}
\right\}\\
&=&\max\left\{
\begin{array}{c}
0,\\
0.83729-0.52595,\\
0.70714-0.52595,\\
0.83729-0.70714\\
\end{array}
\right\}\\
&=&0.31134
\end{eqnarray*}
\begin{eqnarray*}
U&=&\min\left\{
\begin{array}{c}
P(r_a|c),\\
P(r'_{a'}|c),\\
P(r,a|c)+P(r',a'|c),\\
P(r_a|c)-P(r_{a'}|c)+\\+P(r,a'|c)+P(r',a|c)
\end{array}
\right\}\\
&=&\min\left\{
\begin{array}{c}
0.83729,\\
1-0.52595,\\
0.35286+0.20143,\\
0.83729-0.52595+\\+0.35428+0.09143
\end{array}
\right\}\\
&=&0.47405\\
\end{eqnarray*}
Therefore, 
\begin{eqnarray*}
&&W+\sigma L\le f(c) \le W+\sigma U,\\
&&-6.64980+20\times 0.31134 \le f(c)\\
&&\le -6.64980+20\times 0.47405,\\
&&-0.423\le f(c) \le 2.832.\\
\end{eqnarray*}
Then, we apply Theorem \ref{thm1} to the data in Tables \ref{tb1} and \ref{tb2}. $\sigma$ and $W$ are the same as above.\\
And we have,
\begin{eqnarray*}
L&=&\sum_z\max\left\{
\begin{array}{c}
0,\\
P(r_a|z,c)-P(r_{a'}|z,c),\\
P(r|z,c)-P(r_{a'}|z,c),\\
P(r_a|z,c)-P(r|z,c)
\end{array}
\right\}\nonumber\\
&&\times P(z|c)\\
&=&\max\left\{
\begin{array}{c}
0,\\
0.44600-0.05000,\\
0.49010-0.05000,\\
0.44600-0.49010\\
\end{array}
\right\}\times 0.28857\\
&&+\max\left\{
\begin{array}{c}
0,\\
0.99600-0.71900,\\
0.79518-0.71900,\\
0.99600-0.79518\\
\end{array}
\right\}\times 0.71143\\
&=&0.44010\times 0.28857 + 0.27700\times 0.71143\\
&=&0.32407
\end{eqnarray*}
\begin{eqnarray*}
U&=&\sum_z\min\left\{
\begin{array}{c}
P(r_a|z,c),\\
P(r'_{a'}|z,c),\\
P(r,a|z,c)+P(r',a'|z,c),\\
P(r_a|z,c)-P(r_{a'}|z,c)+\\
+P(r,a'|z,c)+P(r',a|z,c)
\end{array}
\right\}\nonumber\\
&&\times P(z|c)\\
&=&\min\left\{
\begin{array}{c}
0.44600,\\
1-0.05000,\\
0.44555+0.20297,\\
0.44600-0.05000+\\+0.04455+0.30693
\end{array}
\right\}\times 0.28857\\
&&+\min\left\{
\begin{array}{c}
0.99600,\\
1-0.71900,\\
0.31526+0.20080,\\
0.99600-0.71900+\\+0.47992+0.00402
\end{array}
\right\}\times 0.71143\\
&=&0.44600\times 0.28857 + 0.28100\times 0.71143\\
&=&0.32862
\end{eqnarray*}
Therefore, 
\begin{eqnarray*}
&&W+\sigma L\le f(c) \le W+\sigma U,\\
&&-6.64980+20\times 0.32407 \le f(c)\\
&&\le -6.64980+20\times 0.32862,\\
&&-0.168\le f(c) \le -0.077.\\
\end{eqnarray*}

\subsubsection{Effective Patients of a Drug}
First, the set $\{C\}$ satisfied the back-door criterion for both $(A,Z)$ and $(A,R)$. By Pearl's adjustment formula, the experimental data needed are:
\begin{eqnarray*}
&&P(r_a|c)=P(r|a,c)=0.66666,\\
&&P(r_{a'}|c)=P(r|a',c)=0.33265,\\
&&P(z_a|c)=P(z|a,c)=0.68878,\\
&&P(z'_{a'}|c)=P(z'|a',c)=0.01232.
\end{eqnarray*}
Then, we apply Li-Pearl's Theorem (Theorem \ref{thm4}) to the data in Table \ref{tb3} and the above experimental data. The benefit vector is $(1,-1,-1,-1)$.\\
We have,
\begin{eqnarray*}
\sigma &=& \beta - \gamma - \theta + \delta\\
&=&1-(-1)-(-1)+(-1)\\
&=&2
\end{eqnarray*}
\begin{eqnarray*}
W&=&(\gamma -\delta)P(r_a|c)+\delta P(r_{a'}|c)+\theta P(r'_{a'}|c)\\
&=&(-1 +1)P(r_a|c)- P(r_{a'}|c)- P(r'_{a'}|c)\\
&=&-1
\end{eqnarray*}
\begin{eqnarray*}
L&=&\max\left\{
\begin{array}{c}
0,\\
P(r_a|c)-P(r_{a'}|c),\\
P(r|c)-P(y_{a'}|c),\\
P(r_a|c)-P(r|c)\\
\end{array}
\right\}\\
&=&\max\left\{
\begin{array}{c}
0,\\
0.66666-0.33265,\\
0.51535-0.33265,\\
0.66666-0.51535\\
\end{array}
\right\}\\
&=&0.33401
\end{eqnarray*}
\begin{eqnarray*}
U&=&\min\left\{
\begin{array}{c}
P(r_a|c),\\
P(r'_{a'}|c),\\
P(r,a|c)+P(r',a'|c),\\
P(r_a|c)-P(r_{a'}|c)+\\+P(r,a'|c)+P(r',a|c)
\end{array}
\right\}\\
&=&\min\left\{
\begin{array}{c}
0.66666,\\
1-0.33265,\\
0.36465+0.30233,\\
0.66666-0.33265+\\+0.15070+0.18232
\end{array}
\right\}\\
&=&0.66666\\
\end{eqnarray*}
Therefore, 
\begin{eqnarray*}
&&W+\sigma L\le f(c) \le W+\sigma U,\\
&&-1+2\times 0.33401 \le f(c)\\
&&\le -1+2\times 0.66666,\\
&&-0.3320\le f(c) \le 0.3333.
\end{eqnarray*}
Then, we apply Theorem \ref{thm2} to the data in Table \ref{tb3} and the above experimental data. $\sigma$, $W$, and $L$ are the same as above.\\
And we have,
\begin{eqnarray*}
U&=&\min\left\{
\begin{array}{c}
P(r_a|c),\\
P(r'_{a'}|c),\\
P(r,a|c)+P(r',a'|c),\\
P(r_a|c)-P(r_{a'}|c)+\\
+P(r,a'|c)+P(r',a|c),\\
\sum_z \sum_{z'} \min\{P(r|z,a,c),\\
P(r'|z',a',c)\}\\
\times \min\{P(z_a|c),P(z'_{a'}|c)\}
\end{array}
\right\}\\
&=&\min\left\{
\begin{array}{c}
0.66666,\\
1-0.33265,\\
0.36465+0.30233,\\
0.66666-0.33265+\\+0.15070+0.18232,\\
\min\{0.92593,0.66944\}\times\\
\min\{0.68878,0.98768\}+\\
\min\{0.92593,0.50000\}\times\\
\min\{0.68878,0.01232\}+\\
\min\{0.09290,0.66944\}\times\\
\min\{0.31122,0.98768\}+\\
\min\{0.09290,0.50000\}\times\\
\min\{0.31122,0.01232\}
\end{array}
\right\}\\
&=&0.49731
\end{eqnarray*}
Therefore, 
\begin{eqnarray*}
&&W+\sigma L\le f(c) \le W+\sigma U,\\
&&-1+2\times 0.33401 \le f(c)\\
&&\le -1+2\times 0.49731,\\
&&-0.3320\le f(c) \le -0.0054.
\end{eqnarray*}

\newpage
\subsection{Distribution Generating Algorithms}
Here, the sample distribution generating algorithms in simulated studies are presented.
\subsubsection{Non-descendant Covariates}
The Algorithm \ref{alg1} is the sample distribution generating algorithm in the simulated study of non-descendant covariates case. It generated both experimental and observational data compatible with Figure \ref{causalg5} ($X,Y,Z$ are binary) that satisfy the general relation provided by Tian and Pearl (i.e., the general relation between experimental and observational data).
\begin{algorithm}[tb]
\caption{Generate sample distributions for non-descendant covariates}
\label{alg1}
\textbf{Input}: $n$, number of sample distributions needed.\\
\textbf{Output}: $n$ sample distributions (observational data and experimental data).
\begin{algorithmic}[1] %[1] enables line numbers
\FOR {$i=1$ to $n$}
    \STATE //$rand(0,1)$ is the function that random uniformly generate a number from $0$ to $1$.
    \STATE // $t_1,t_2,t_3$, and $t_4$ can be interpreted as the number of individuals such that $x\land z$, $x'\land z$, $x\land z'$, and $x'\land z'$ respectively.
    \STATE $t_1=rand(0,1)\times 1000$;
    \STATE $t_2=rand(0,1)\times (1000-t_1)$;
    \STATE $t_3=rand(0,1)\times (1000-t_1-t_2)$;
    \STATE $t_4=$ $1000-t_1-t_2-t_3$;
    \STATE // $o_1,o_2,o_3$, and $o_4$ can be interpreted as the number of individuals such that $x\land z \land y$, $x'\land z \land y$, $x\land z'\land y$, and $x'\land z'\land y$ respectively.
    \STATE $o_1=rand(0,1)\times t_1$;
    \STATE $o_2=rand(0,1)\times t_2$;
    \STATE $o_3=rand(0,1)\times t_3$;
    \STATE $o_4=rand(0,1)\times t_4$;
    \STATE // Each $c_i$ corresponding to a sample distribution.
    \STATE // The following are experimental data that satisfied the general bounds provided by Tian and Pearl.
    \STATE $P(y|do(x),z,c_i)=rand(0,1)\times \frac{t_2}{t_1+t_2}+\frac{o_1}{t_1+t_2}$;
    \STATE $P(y|do(x'),z,c_i)=rand(0,1)\times \frac{t_1}{t_1+t_2}+\frac{o_2}{t_1+t_2}$;
    \STATE $P(y|do(x),z',c_i)=rand(0,1)\times \frac{t_4}{t_3+t_4}+\frac{o_3}{t_3+t_4}$;
    \STATE $P(y|do(x'),z',c_i)=rand(0,1)\times \frac{t_3}{t_3+t_4}+\frac{o_4}{t_3+t_4}$;
    \STATE // The following are observational data.
    \STATE $P(x,y,z|c_i)=o_1/1000$;
    \STATE $P(x,y,z'|c_i)=o_3/1000$;
    \STATE $P(x,y',z|c_i)=(t_1-o_1)/1000$;
    \STATE $P(x,y',z'|c_i)=(t_3-o_3)/1000$;
    \STATE $P(x',y,z|c_i)=o_2/1000$;
    \STATE $P(x',y,z'|c_i)=o_4/1000$;
    \STATE $P(x',y',z|c_i)=(t_2-o_2)/1000$;
    \STATE $P(x',y',z'|c_i)=(t_4-o_4)/1000$;
\ENDFOR
\end{algorithmic}
\end{algorithm}
\subsubsection{Partial Mediators}
The observational data compatible with Figure \ref{causalg2} ($X,Y,Z$ are binary) in the simulated study of partial mediators case was generated by Algorithm \ref{alg2}. The experimental data needed was computed via adjustment formula because the set $\{C\}$ satisfied the back-door criterion for both $(X,Z)$ and $(X,Y)$.
\begin{algorithm}[tb]
\caption{Generate sample distributions for partial mediators}
\label{alg2}
\textbf{Input}: $n$, number of sample distributions needed.\\
\textbf{Output}: $n$ sample distributions (observational data in conditional probability tables).
\begin{algorithmic}[1] %[1] enables line numbers
\FOR {$i=1$ to $n$}
    \STATE //$rand(0,1)$ is the function that random uniformly generate a number from $0$ to $1$.
    \STATE // Each $c_i$ corresponding to a sample distribution.
    \STATE $P(x|c_i)=rand(0,1)$;
    \STATE $P(z|x,c_i)=rand(0,1)$;
    \STATE $P(z|x',c_i)=rand(0,1)$;
    \STATE $P(y|x,z,c_i)=rand(0,1)$;
    \STATE $P(y|x',z,c_i)=rand(0,1)$;
    \STATE $P(y|x,z',c_i)=rand(0,1)$;
    \STATE $P(y|x',z',c_i)=rand(0,1)$;
\ENDFOR
\end{algorithmic}
\end{algorithm}
\subsubsection{Pure Mediators}
The observational data compatible with Figure \ref{causalg3} ($X,Y,Z$ are binary) in the simulated study of pure mediators case was generated by Algorithm \ref{alg3}. The experimental data needed was computed via adjustment formula because the set $\{C\}$ satisfied the back-door criterion for $(X,Y)$.
\begin{algorithm}[tb]
\caption{Generate sample distributions for pure mediators}
\label{alg3}
\textbf{Input}: $n$, number of sample distributions needed.\\
\textbf{Output}: $n$ sample distributions (observational data in conditional probability tables).
\begin{algorithmic}[1] %[1] enables line numbers
\FOR {$i=1$ to $n$}
    \STATE //$rand(0,1)$ is the function that random uniformly generate a number from $0$ to $1$.
    \STATE // Each $c_i$ corresponding to a sample distribution.
    \STATE $P(x|c_i)=rand(0,1)$;
    \STATE $P(z|x,c_i)=rand(0,1)$;
    \STATE $P(z|x',c_i)=rand(0,1)$;
    \STATE $P(y|z,c_i)=rand(0,1)$;
    \STATE $P(y|z',c_i)=rand(0,1)$;
\ENDFOR
\end{algorithmic}
\end{algorithm}

\end{document}